\documentclass[a4paper,11pt]{ijamas}

\usepackage[pdftex]{graphicx}
\usepackage{algorithm}
\usepackage{algorithmic}
\usepackage[squaren,Gray]{SIunits}

\begin{document}


\title{An Algorithm and Heuristic based on Normalized Mutual Information  for
Dimensionality Reduction and Classification of Hyperspectral images}

\author{\textbf{Elkebir Sarhrouni$^1$ , Ahmed Hammouch$^2$ and Driss Aboutajdine$^1$}}

\date{$^1$LRIT,FSR, UMV-A\\
Rabat, Morocco\\
sarhrouni436@yahoo.fr\\[0.3cm]
$^2$LGGE, ENSET, UMV-SOUISSI\\
Rabat, Morocco\\}

\maketitle


\begin{abstract}
\noindent \emph{In the feature classification domain, the choice of data affects widely the results.
The Hyperspectral image (HSI), is a set of more than a hundred bidirectional measures
(called bands), of the same region (called ground truth map: GT). The HSI is modelized
at a set of N vectors. So we have N features (or attributes)  expressing  N vectors of measures
for C substances (called classes). The problematic is that it's pratically impossible to investgate
all possible subsets. So we must find K vectors among N, such as relevant and no redundant
ones; in order to classify substances. Here we introduce an algorithm based on Normalized
Mutual Information to select relevant and no redundant bands, necessary to increase
classification accuracy of HSI.}

\medskip

\noindent\textbf{Keywords:} Feature Selection, Normalized Mutual information, Hyperspectral images, Classification,  Redundancy.

\medskip

\noindent\textbf{Mathematics Subject Classification:} 68U10, 68R05.\\
\noindent\textbf{Computing Classification System:} I.4.7, I.4.8, I.4.9.

\end{abstract}


\section{Introduction}

The Hyperspectral image AVIRIS 92AV3C, Airborne Visible Infrared Imaging Spectrometer, \cite{[2]} is a substitution of 220 images for the region "Indiana Pine" at "north-western Indiana", USA \cite{[1]}. The 220 called bands are taken between \unit{0.4}{\micro}m and \unit{2.5}{\micro}m. Each band has 145 lines and 145 columns. The ground truth map is also provided, but only 10366 pixels (49\%) are labeled fro 1 to 16. Each label indicates one from 16 classes. The zeros indicate pixels how are not classified yet, see Figure.1.
\begin{figure}[!th]
\centering
\includegraphics[width=4.5in]{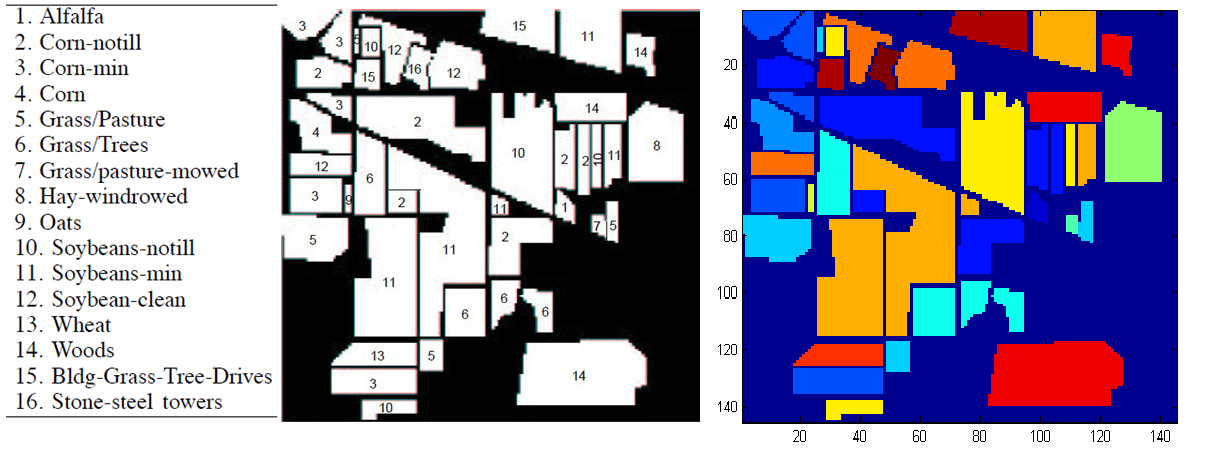}
\caption{The Ground Truth map of AVIRIS 92AV3C and the 16 classes }
\end{figure}
The hyperspectral image AVIRIS 92AV3C contains numbers between 955 and 9406. Each pixel of the ground truth map has a set of 220 numbers (measures) along the hyperspectral image. This numbers (measures) represent the reflectance of the pixel in each band. So the pixel is shown as vector off 220 components,see Figure .2.\\
\begin{figure}[!h]
\centering
\includegraphics[width=4.5in]{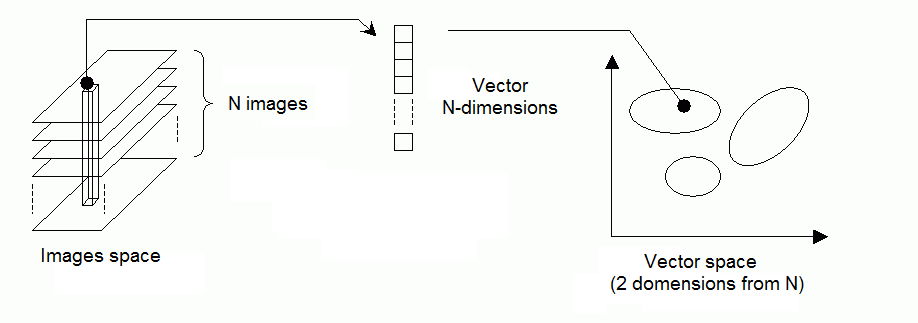}
\caption{The notion of  pixel vector }
\end{figure}
To classify pixels of HSI, we find some problems realed to their high dimentions, that needs many cases to detect the relationship between the vectors and the classes, according to Hughes phenomenon \cite{[10]}. other problems are related to the redundant images (bands); they complicate the learning system and product incorrect prediction \cite{[14]}. So we must pick up only the relevant bands. Now we can identfy the problematic relaited to HSI as a dimentionality reduction. It is commonly reencountered when we have $N$ features (or attributes) that express $N$ vectors of measures for $C$ substances (called classes). The problematic is to find $K$ vectors among $N$, such as relevant and no redundant ones; in order to classify substances. The number of selected vectors $K$  must be lower than $N$ regarding the problems above. So we must choose relevant vectors, that means there ability to predicate the classes. Indeed the bands don’t all contain the information; some bands are irrelevant like those affected by various atmospheric effects, see Figure.7, we can show the atmospheric effects on bands: 155, 220 and other bands; so the classification accuracy decreases. We can reduce the dimensionality of hyperspectral images by selecting only the relevant bands (feature selection or subset selection methodology), or extracting, from the original bands, new bands containing the maximal information about the classes, using any functions, logical or numerical (feature extraction methodology) \cite{[8],[9],[11]} . Here we propose an algorithm based on mutual information, and normalised mutual information fo reducing dimensionality. This will be as bellow: pick up the relevant bands first, and avoiding redundancy second. We illustrate the principea of this algorithm using synthetic bands for the scene of HIS AVIRIS 92AV3C\cite{[1]} , see Figure.5.  Then we validate its effect by applying it to real datat of HSI AVIRIS 92AV3C. Here each pixel is shown as a vector of  220 components. Figure.2 shows the vector pixels notion \cite{[7]}. Reducing dimensionality means selecting only the dimensions caring a lot of information regarding the ground truth map (GT).\\
In the literature, we can cite some methods related to the dimension reduction; but this action is accompanied by transformation of the multivariate data. Well-known methods include  principal component analysis, factor analysis, projection pursuit, independent component analysis (ICA).
Principal Component Analysis, or PCA \cite{[26]}, is widely used in signal processing, statistics, and neural computing \cite{[23],[27],[28]}. The PCA search another space with lower diemnsion, and  when the data will be clearly separated. 
The independent component analysis ICA search also a space with  lower diemntion, and which the sources of data are separated \cite{[28],[29],[25]}; so which desired action is minimizing the statistical dependence of the components and the components didn't necessary orthogonal. Now due to the dimentionality of Hyperspectral Image, the question proposed is: can we find some principal components, or independant component  among the 220 bands, without transformation of data? The response is to colculate the separation matrix using normalized mutual information as a cost fonction. We cite here that some works  \cite{[29],[30],[31],[32]} had already use the mutual information ( but not normalized form of MI) and neural network, to separate blinded sources.


\section{ Principle of  Feature Selection with Mutual Informationl}
\subsection{Definition of Mutual Information}
The MI is a measure of information contained in both tow ensembles of random variables A and B:
\[
I(A,B)=\sum\;p(A,B)\;log_2\;\frac{p(A,B)}{p(A).p(B)}
\]

Let us consider the ground truth map (GT), and bands as ensembles of random variables, the MI between GT and each band calculates their interdependence. Fano \cite{[14]} has demonstrated that as soon as mutual information of already selected features $X$ and the classe $C$ has high value, the error probability of classification decreases, according to the formula bellow:
\[\;\frac{H(C/X)-1}{Log_2(N_c)}\leq\;P_e\leq\frac{H(C/X)}{Log_2}\; \]with :
\[
\;\frac{H(C/X)-1}{Log_2(N_c)}=\frac{H(C)-I(C;X)-1}{Log_2(N_c)}\; \] and :
\[    P_e\leq\frac{H(C)-I(C;X)}{Log_2}=\frac{H(C/X)}{Log_2}\; \]
Here the conditional entropy $H(C/X)$ is calculated between the ground truth map (i.e. the classes $C$) and the subset of bands candidates $X$. $ N_{c}$ is the number of classes. So when the features $X$ have a higher value of mutual information with the ground truth map, (is more near to the ground truth map), the error probability will be lower. But it's impractical to compute this conjoint mutual information $I(C;X)$, regarding the high dimensionality \cite{[14]}.\\
Figure.6. shows the MI between the GT and synthetic bands. The figure.7 indicates the MI between the GT and the real bands of HIS AVIRIS 92AV3C \cite{[1]}.\\
Some studies use a thresholding to choice the more informative bands. Guo \cite{[3]} uses the mutual information to select the top ranking band, and a filter based algorithm to decide if there neighbours are redundant or not. Sarhrouni \cite{[17]} use also a filter strategy based algorithm on MI to pick up relevant bands. A wrapper strategy based algorithm on MI, Sarhrouni \cite{[18]} is also introduced.\\
By a thresholding, for example with a threshold 0.4, see Figure.6, we eliminate the no informative bands: $A_3$, $A_7$ and $A_9$. With other threshold, we can retain fewer bands. We can visually show this effectiveness of MI to choice relevant features in Figure.6, and Figure.7.
\subsection{Normalized Mutual information}
In our study we use two forms of normalized mutual information, and we compare their results.
\subsubsection{First Form}
The firrst form of Mutual Information is defined as bellow: 

\[AS(A,B)=\frac{MI(A,B)}{H(A)} ;          \] and \[AS(B,A)=\frac{MI(B,A)}{H(B)}\]

$H(X)$ is the Entropy of set random variable $X$. This is an $asymetric$ measure of MI.\\
Inspecting this formula We can make this observation:
\textit{"When $AS(A,B)$ is near to 1 i.e. 100\%, it means that $A$ exchange all its information with $B$."}\\
So we can use this as a $measure$ of $redundancy$.
 Numerous studies use Normalized Mutual Information for recalling images etc.\cite{[20],[21],[22]}.\\
Figure.3 illustrates the MI and Normalized MI.
\begin{figure}[!th]
\centering
\includegraphics[width=3.0in]{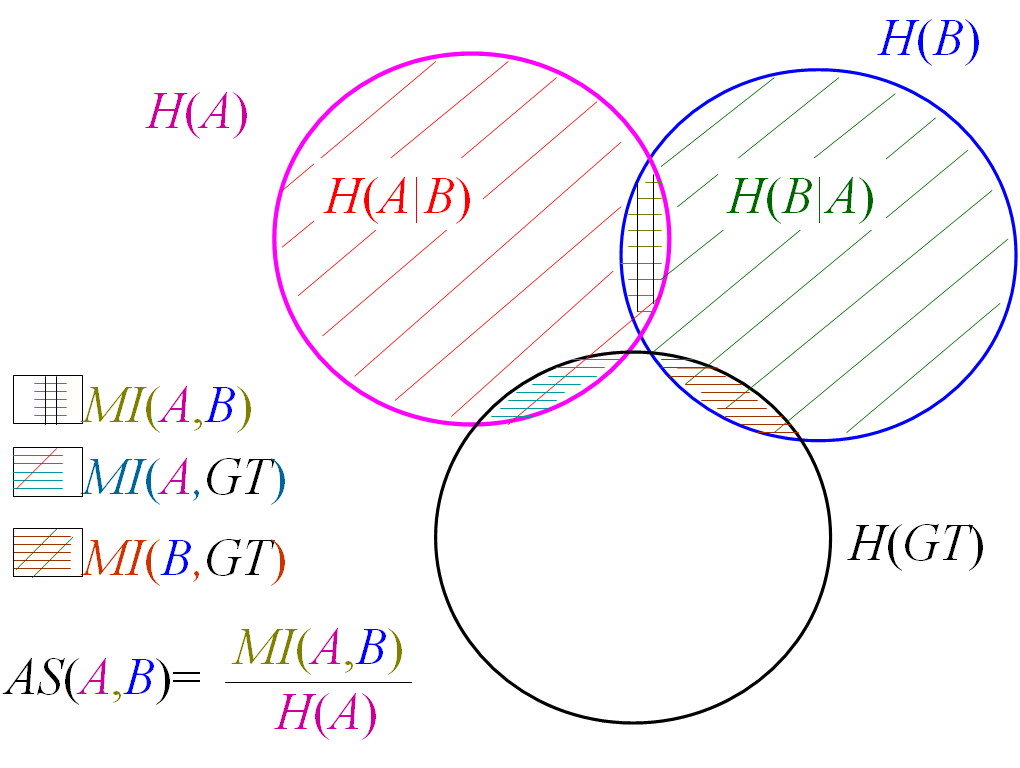} 
\caption{Normatized Mutual Information}
\end{figure}
\subsubsection{Second Form}
This is one of normalized form of Mutual Information  introduced by \cite{[19]}. It's defined as bellow: 
\[
U(A,B)=\frac{MI(A,B)}{\sqrt[2](H(A).H(B))}
\]
$H(X)$ is the Entropy of set random variable $X$.\\
Figure.4 shows that $ normalized$ $mutual$ $information$ means how much information is partaged between $A$ and $B$ relatively at all information contained in both $A$ and $B$.
\begin{figure}[!th]
\centering
\includegraphics[width=3.0in]{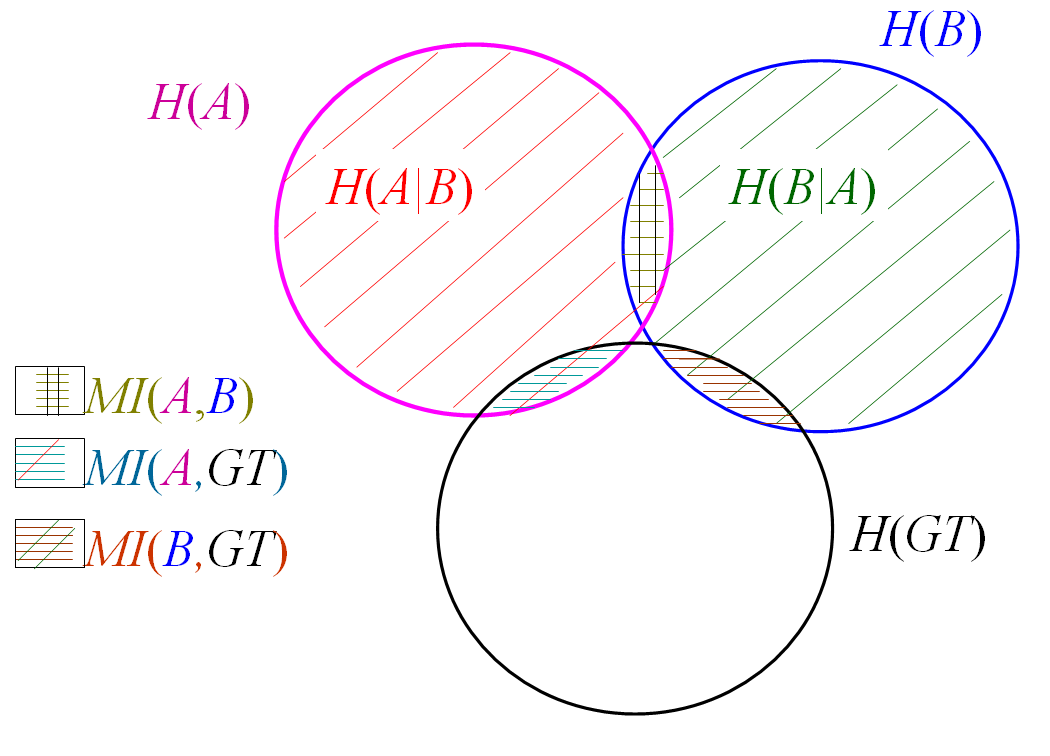} 
\caption{Illustration of Symetric Uncertainty}
\end{figure}
\section{Principle of the Proposed Method and Algorithm}
For this section we use 19 synthetic bands from the GT, Figure.1, by adding noise, cutting some substances etc. see Figure.5. Each band has 145 lines and 145 columns. Only 10366 pixels are labelled from 1 to 16. Each label indicates one from 16 classes. The zeros indicate pixels how are not classified yet, Figure.1. We can show the Mutual information of GT and the synthetic bands at Figure.6.
\begin{figure}[!th]
\centering
\includegraphics[width=3.50in]{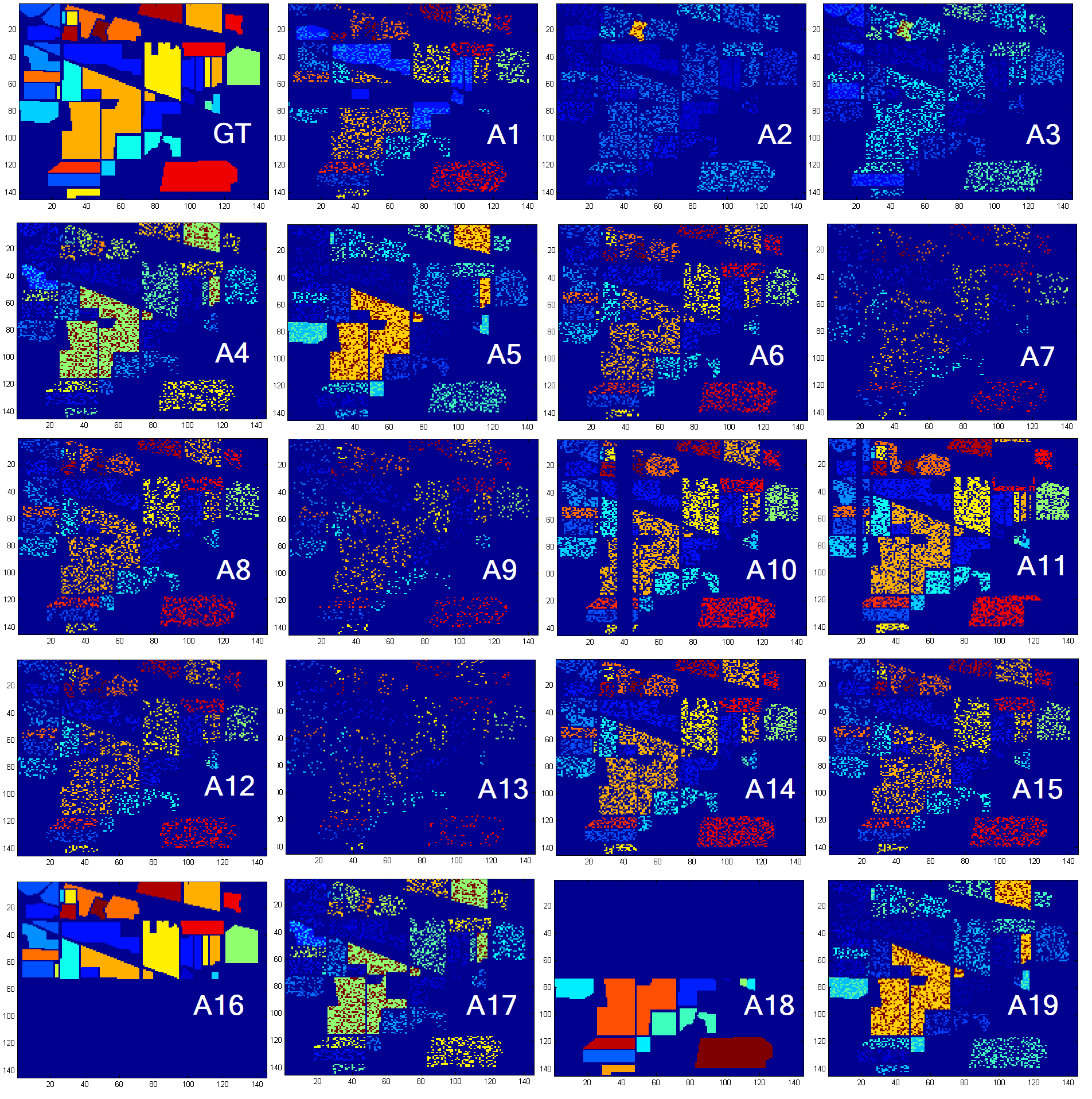}
\caption{The synthetic bands used for the study .}
\end{figure}

\begin{figure}[!th]
\centering
\includegraphics[width=3.5in]{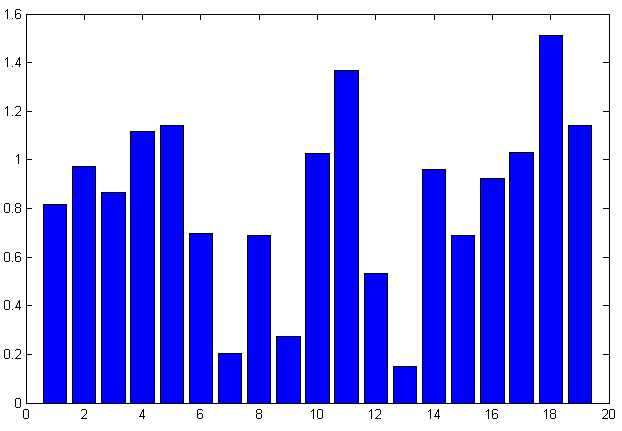}
\caption{Mutual Information of GT and synthetic bands   .}
\end{figure}
\subsection{Selection of  relevant bands}
With a threshold 0.4 of MI calculated in Figure.4 we obtain 16 relevant bands $ A_i$  with :\\
\textit{ i=\{1,2,3,4,5,6,8,10,11,12,14,15,16,17,18,19\}}.\\
 We can visually verify the resemblance of GT and the bands more informative, both in synthetic and the real data bands of AVIRIS 92AV3C. See Figure.6 and Figure.7.

\subsection{Detection of no Redundant Bands}
First: We order the remaining bands, in increasing order of there MI with the GT. So we have:\\
$\{A_{12} A_8 A_{15 }A_6 A_1 A_3 A_{14 }A_{16} A_2 A_{10 }A_{17 }A_4 A_{19} A_5 A_ {11} A_ {18}\}$\\

Second: We fixe a threshold to control redundancy, here 0.7. Then we compute the Normalised  $AS (A_{i},A_{j})$ MI (Table 1), and the $Symmetric$ $Uncertainty$   $U (A_{i},A_{j})$ (Table 2): for all couple $( i, j )$ of the ensemble:\\
 $S=\{8,15,6,1,3,14,16,2,10,17,4,19,5,11,18\}$. \\

\begin{table}[!h]
\center
\caption{THE NORMALIZED MI OF THE RELEVANT SYNTHETIC BANDS. }
\includegraphics[width=5.0in]{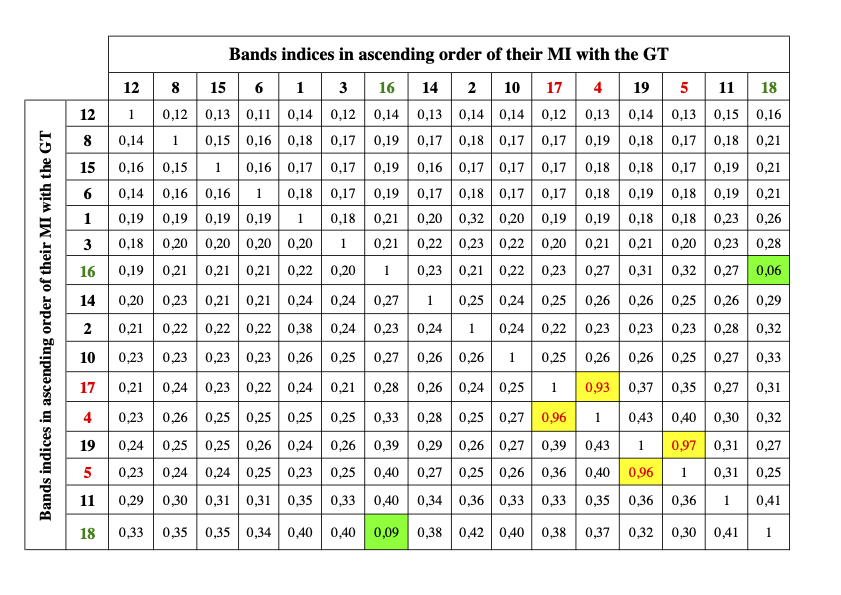}
\end{table}
\begin{table}[!h]

\center
\caption{THE SYMMETRIC UNCERTAINTY OF THE RELEVANT SYNTHETIC BANDS. }
\includegraphics[width=4.50in]{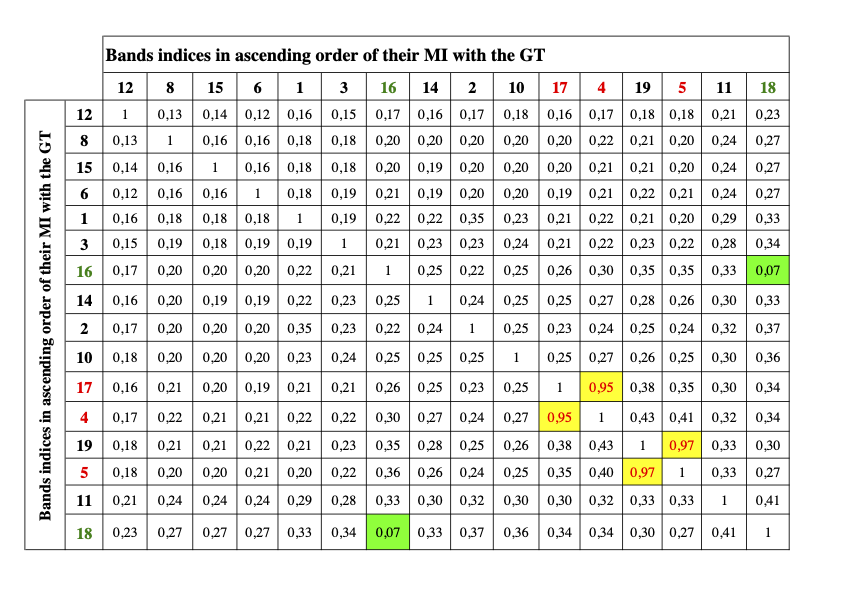}
\end{table}
Observation 1 : Figure.5 shows that the band $A_{17}$ is practically the same at  $A_{4}$.  Table 1  shows that $AS (A_{17}, A_{4})$ and $AS (A_{4}, A_{17})$ near to 100\% (respectively 0.96 and 0.93). It means that both $A_{17}$ an $A_{4}$ share their information with each other. So this indicates a high redundancy. (Table 2 allows the same observation)\\

Observation 2 : Figure.5 shows that the bands $ A_{16}$ and $A_{18}$ are practically disjoint, i.e. they are not redundant. Table.1. shows $AS(A_{16}, A_{18})=0.06$  and $AS(A_{18}, A_{16})=0.09$. It means that no information is shared between $A_{16}$ and $A_{18}$ .So this indicates no redundancy  (Table 2 allows the same observation).\\
This makes an interest result : the ensemble of selected bands became $SS= \{16, 18\}$. $A_{16}, A_{18}$ will be discarded from the Table .1.  (The same for Table 2 )\\
 Now we can emit this rule:
 
\textit{" Each band candidate will be added to the ensemble of already selected ones, SS, if and only if their Normalized Mutual Information values with all elements of SS, are less than the thresholds (here 0.7)."}\\
Algorithm 1 implements this rule.
\begin{algorithm}  
\vspace{0.20cm}                    
\caption{: $Band$ is the HSI. Let $Th$$_{relevance}$ the threshold for selecting bands more informative, $Th$$_{redundancy}$  the threshold for redundancy control. }  
\center        
\label{algo 1}  
\begin{algorithmic}
\STATE 1) Compute the Mutual Information $(MI)$ of the bands and the Ground Truth map.
\STATE 2) Make bands in ascending order by there $MI$ value 
\STATE 3) Cut the bands that have a lower value than the threshold $Th$$_{relevance}$, the subset remaining is $S$.
\STATE 4) Initialization:  $n=length(S), i=1$,  $D$ is a bidirectional array values=1;\\
//any value greater than 1 can be used, it's useful in step 6)
\STATE 5) Computation of bidirectional Data $D(n,n)$:
\FOR{  1:=1  to n step 1 }
\FOR{	 j:=1  to n step 1 }
\STATE $D(i,j)= AS(Band_{S(i)},Band_{S(j)});$  // $with$ $AS(A,B)=\frac{MI(A,B)}{H(A)}$
\STATE	//$Or$ $D(i,j)= U(Band_{S(i)},Band_{S(j)});$  $with$ $U(A,B)=\frac{MI(A,B)}{\sqrt[2](H(A).H(B))}$
\ENDFOR
\ENDFOR  
\STATE 6) $SS=\{\}$ ; // $Initialization$ $of$ $the$ $Output$ $of$ $the$ $algorithm$
\WHILE{$ min(D)<Th_{redundancy}$}
\STATE 	// $Pick$ $up$ $the$ $argument$ $of$ $the$ $minimum$ $of$ $D$
\STATE $(x,y)= argmin(D(.,.));$     
\IF {$ \forall \  l \in SS\, \  D(x,l)<Th_{redundancy}$}  
\STATE	 // $check$ $x$ $ is$ $ not$ $ redundant$ $ with$ $ the$ $ already$ $ selected$ $ bands$
 \STATE $SS=SS \cup \{x\}$   
\ENDIF
\STATE $D(x,y)=1$;  // $The$ $ cells$ $ D(x,y)$  $will$ $ not$ $ be$ $ checked$ $ as$ $ minimum$ $ again	$
\ENDWHILE
\STATE 7)  Finish: The  final subset  $SS$ contains bands according to the the couple of  thresholds \ ( $Th_{relevance}$,$T_{redundancy}$). 
\end{algorithmic}
\end{algorithm}

\section{Application On HIS AVIRIS 92AV3C  }
The Agorithm.1 implement the proposed method .
Here we use real data, hyperspectral image AVIRIS 92AV3C \cite{[1]}, as input of the algorithm.1. 50\% of the labelled pixels are randomly chosen and used in training; and the other 50\% are used for testing classification \cite{[3],[17],[18]}. The classifier used is the SVM \cite{[5],[12],[4]}.
The Agorithm.1 implement the proposed method .

\subsection{Mutual Information Curve of Bands}
We must eliminate no informative bands from the remaining subset bay thresholding, see the proposed algorithm. Figure.7 gives the MI of the HSI AVIRIS 92AV3C with the ground truth GT.

\begin{figure}[!h]
\centering
\includegraphics[width=3.5in]{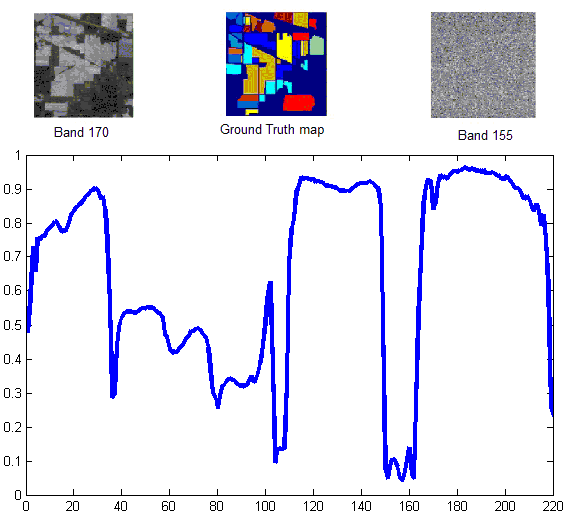}
\caption{Mutual information of GT and AVIRIS bands   .}
\end{figure}

\subsection{Results}
From the remaining subset bands, we must eliminate redundant ones using the proposed algorithm. Table 3 an Table 4 give the accuracy off classification for a number of bands with several thresholds, for the two forms of normalized MI.

\subsection{Discussion}

Results in Table.3 and Table 4 allow us to distinguish six zones of couple values of thresholds $(TH,IM)$:\\
\begin{table}[!th]
\center
\caption{Classification Accuracy for several couples of thresholds (TH,IM) and their corresponding number of bands retained, (First Form of Normalized MI). }
\includegraphics[width=6.60in]{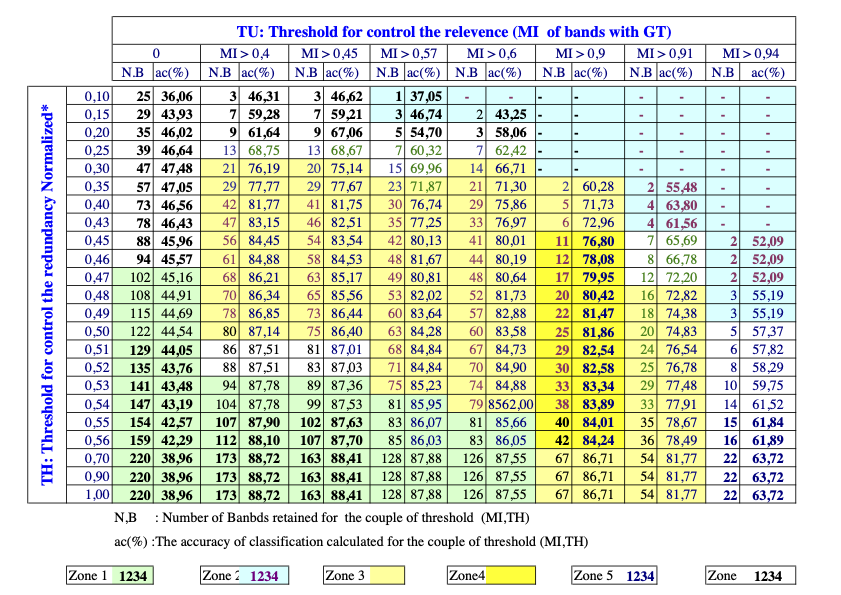}
\end{table}

\begin{table}[!th]
\center
\caption{Classification Accuracy for several couples of thresholds (TH,IM) and their corresponding number of bands retained, (Second Form of Normalized MI). }
\includegraphics[width=6.8in]{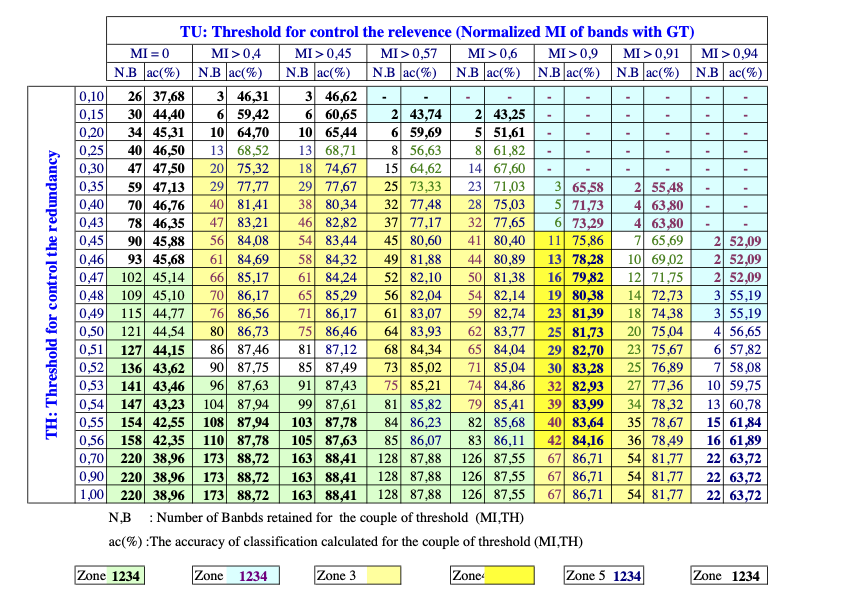}
\end{table}

First case: lower values of TU and higher values of TH, this is practically no control of relevance and no control of redundancy. So there is no action of the algorithm (Zone1).\\

Second case: Higher values TU and lower values of TH, this is a hard selection: a few more relevant and no redundant bands are selected (Zone2). \\

THired case (Zone3): This is an interesting zone. We can have easily 80\% of classification accuracy with about 41 bands.\\

Fourth case (Zone4): This is the very important zone; we have the very useful behaviours of the algorithm. For example with a few numbers of bands 17 we have classification accuracy 80\%.\\

Fifth case: Here we make a hard control of redundancy, but the bands candidates are more near to the GT, and they my be  more redundant. So we can’t have interesting results.  (Zone5)\\ 

Sixth case (Zone6): When we do not control properly the relevance, some bands affected bay transfer affects may be non redundant, and can be selected, so the accuracy of classification is decreasing. \\

 We conclude that this algorithm is very effectiveness for redundancy and relevance control, in feature selection area. \\

 The most difference of this heuristic regarding previous works is the separation of the operations: avoiding redundancy and selecting more informative bands. Sarhrouni \cite{[17]} use also a filter strategy an MI based algorithm to select bands, and an another wrapper strategy algorithm also based on MI \cite{[18]}. Guo \cite{[3]} used a filter strategy with threshold control redundancy, but in those works, the tow process, i.e avoiding redundancy and avoiding no informative bands, are made at the same time by the same threshold.\\
Figure.8 illustrates the reconstruction of the ground truth map GT, for the first form of normalised MI. The redundancy threshold 0.56 and relevance threshold IM=0.9. The accuracy classification is 84.24\% for 42 bands selected. The generalisation of classification to the entire scene Indiana Pin \cite{[1]} illustrates the power of the proposed method: the pixels not labelled in GT, are here classified. 
\begin{figure}[!th]
\centering
\includegraphics[width=5.0in]{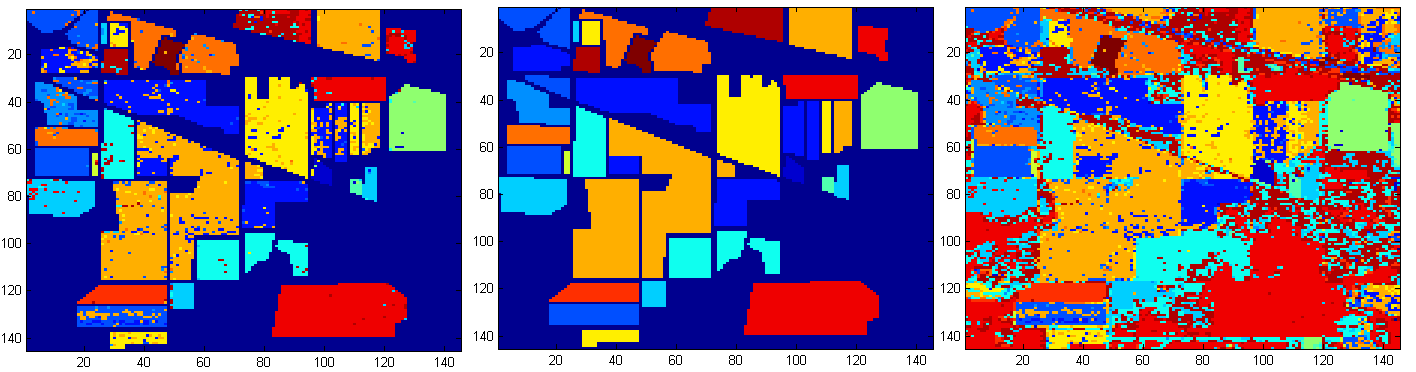} 
\caption{In the middle the GT of AVIRIS 92AV3C. In the left: Reconstructed Truth map (GT) with the proposed algorithm for TH=0.56 and MI=0.9; the  accuracy = 84.24\% for only 42 bands. In right the generalization of classification for all Indiana Pine regions.}
\end{figure}\\
Figure.9 also, illustrates the reconstruction of the ground truth map GT, for the second form of normalised MI. The redundancy threshold 0.56 and relevance threshold IM=0.9. The accuracy classification is 84.16\% for 42 bands selected. The generalisation of classification to the entire scene Indiana Pin \cite{[1]} illustrates the power of the proposed method: the pixels not labelled in GT, are here classified. \\
\begin{figure}[!th]
\centering
\includegraphics[width=5.0in]{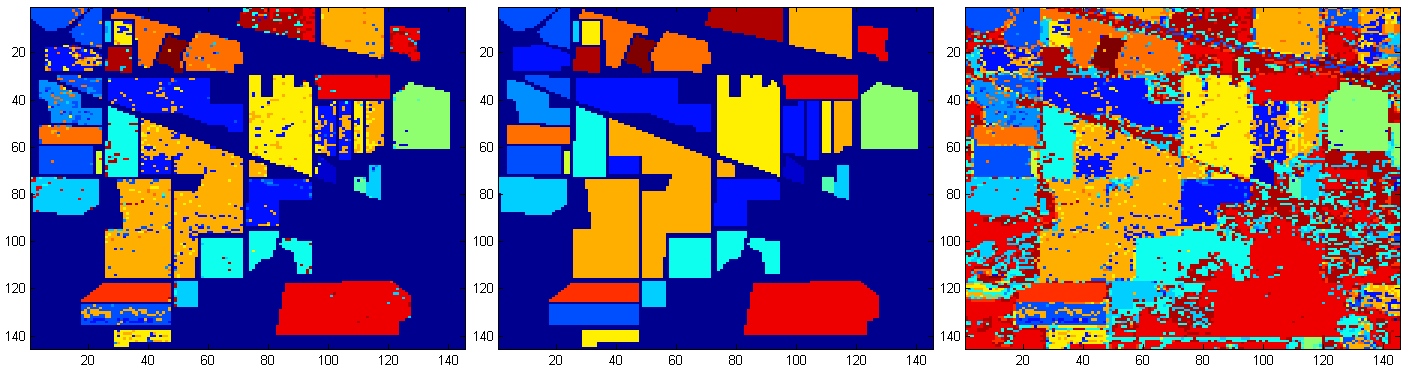} 
\caption{In the middle the GT of AVIRIS 92AV3C. In the left: Reconstructed Truth map (GT) with the proposed algorithm for TH=0.56 and MI=0.9; the  accuracy = 84.24\% for only 42 bands. In right the generalization of classification for all Indiana Pine regions.}
\end{figure}\\
Now, we can view that the two forms of normalized MI give practically the same performances. But Table 2 is a symmetric matrix; and affects largely the complexity of the algorithm. 

\section{Conclusion}
The features selection in high dimensionality is problematic that is always open.The combinatory search and test for all possible subset, is pratically impossible. We can use some heuristic methods and algorithms  to select no optimal subset features, but useful in practical field. The subset retained must be relevant and no redundant . In this paper we introduce an heuristic in order to process separately the relevance and the redundancy. We apply our method to classify the region Indiana Pin with the Hyperspectral Image AVIRIS 92AV3C. This algorithm is a Filter strategy (i.e. with no call to classifier during the selection). In the first step, by thresholding we use mutual information to pick up relevant bands (like most method already used). The second step uses Normalized Mutual information to measure redundancy. We conclude the effectiveness of our method and algorithm to select the relevant and no redundant bands. This algorithm allows us a wide area of possible fasted applications. But no guaranties that the chosen bands are the optimal ones; and some times redundancy can be important to reinforcement of learning classification system. So the thresholds controlling relevance redundancy is a very useful tool to calibre the selection, in real time applications, regarding that in commercial applications, the inexpensive filtering algorithms are urgently preferred.

\bibliographystyle{dcu}
\bibliography{IJAMAS.bib}

@Book{[1], 
  author = 	 {D. Landgrebe},
  title = 	 {On information extraction principles for hyperspectral data: A white paper},
  publisher = 	 {Purdue University, West Lafayette, IN,  Technical Report, School of Electrical and Computer Engineering},
  year = 	 {1997},
  address = 	 {Téléchargeable ici :  http://dynamo.ecn.purdue.edu/~landgreb/whitepaper.pdf. }}

@Book{[2], 
  author = 	 {Perdue},
  title = 	 {},
  publisher = 	 {document avaible at: ftp://ftp.ecn.purdue.edu/biehl/MultiSpec/},
  year = 	 {97},
  address = 	 {}

 %Guo:2006,

@Article{[3],
  author = 	 {B. Guo and Steve R. Gunn and R. I. Damper Senio  and J. D. B. Nelson},
  title = 	 {Band Selection for Hyperspectral Image Classification Using Mutual Information},
  journal = 	 {IEE GEOSCIENCE AND REMOTE SINSING LETTERS},
  year = 	 { 2006},
  volume = 	 {Vol 3},
  number = 	 {4},
  pages = 	 {}}

@Article{[4], 
  author = 	 {B. Guo and Steve R. Gunn and R. I. Damper Senio  and J. D. B. Nelson.},
  title = 	 {Customizing Kernel Functions for SVM-Based Hyperspectral Image Classification},
  journal = 	 {IEEE TRANSACTIONS ON IMAGE PROCESSING},
  year = 	 {2008},
  volume = 	 {Vol 17},
  number = 	 {4},
  pages = 	 {}}

@article{[5], 
 author = {Chang, Chih-Chung and Lin, Chih-Jen},
  title = {{LIBSVM}: A library for support vector machines},
 journal = {ACM Transactions on Intelligent Systems and Technology},
 volume = {2},
 issue = {3},
 year = {2011},
 pages = {27:1--27:27},
 note =	 {Software available at \url{http://www.csie.ntu.edu.tw/~cjlin/libsvm}}

%Nathalie:2009,

@Book{[7],
  author = 	 {David Kern\'eis},
  title = 	 {Am\'elioration de la classification automatique des fonds marins par la fusion multicapteurs acoustiques},
  publisher = 	 { PhD thesis, ENST BRETAGNE, universit\'e de Rennes },
  year = 	 {2007},
  note =	{Chapitre3, R\'eduction de dimensionalit\'e et classification, Page.48}}

@Article{[8], 
  author = 	 {N. Kwak and C. Choi},
  title = 	 {Featutre extraction based on direct calculation of mutual information},
  journal = 	 {IJPRAI},
  year = 	 { 2007},
  volume = 	 {Vol 21},
  number = 	 {7},
  pages = 	 {1213--1231}}

@Article{[9], 
  author = 	 {Nojun Kwak and C. Kim},
  title = 	 {Dimensionality Reduction Based on ICA Regression Problem},
  journal = 	 {Lecture Notes in Computer Science},
  year = 	 {2006},
  volume = 	 {1431},
  number = 	 {},
  pages = 	 {1--10}}

@Article{[10],
  author = 	 { G. Huges},
  title = 	 {On the mean accuracy of statistical pattern recognizers},
  journal = 	 {Information Thaory, IEEE Transactionon},
  year = 	 {1968},
  volume = 	 {Vol 14},
  number = 	 {1},
  pages = 	 {55--63}}

@Article{[11], 
  author = 	 {YANG and Yiming and Jan O. PEDERSEN.},
  title = 	 {A comparative study of feature selection in text categorization},
  journal = 	 {Proceedings of the Fourteenth International Conference on Machine Learning(ICML'97)},
  year = 	 {1997},
  publisher=    {San Francisco, CA, USA:Morgan Kaufmann Publishers Inc},
  volume = 	 {5},
  number = 	 {},
  pages = 	 {412--420}}

@article{[12], 
 author = {Chih-Wein Hsu and Chih-Jen  Lin},
title = {A comparison of methods for multiclass support vector machines },
 journal = {Neural Networks, IEEE Transactions on },
 volume = {13},
 issue = {2},
 year = {2002},
 pages = {415--425},
 note =	 {Sponsored by: IEEE Computational Intelligence Society}}

@Article{[14], 
  author = 	 {Yu Lei  and   Liu Huan},
  title = 	 {Efficient Feature Selection via Analysis of Relevance and Redundancy},
  journal = 	 {Journal of Machine Learning Research},
  year = 	 {2004},
  volume = 	 {5},
  number = 	 {},
  pages = 	 {1205--1224}}

@Article{[17], 
  author = 	 {E. Sarhrouni and A. Hammouch and D. Aboutajdine},
  title = 	 {Dimensionality reduction and classification feature using mutual information applied to hyperspectral images: a filter strategy based algorithm},
  journal = 	 { Appl. Math. Sci},
  year = 	 {2012},
  volume = 	 {6},
  number = 	 {101-104},
  pages = 	 {5085--5095}}

@Article{[18], 
  author = 	 {E. Sarhrouni and A. Hammouch and D. Aboutajdine},
  title = 	 {Dimensionality reduction and classification feature using mutual information applied to hyperspectral images: a wrapper strategy algorithm based on minimizing the error probability using the inequality of Fano},
  journal = 	 { Appl. Math. Sci},
  year = 	 {2012},
  volume = 	 {6},
  number = 	 {101-104},
  pages = 	 {5073--5084}}

@book {[19], 
    AUTHOR = {Witten,  Eibe Frank},
     TITLE = {Data Mining: Practical Machine Learning Tools and Techniques},
    SERIES = {},
    VOLUME = {},
 PUBLISHER = {Morgan Kaufmann,2 edition},
   ADDRESS = {San Francisco},
      YEAR = {2005},
     PAGES = {},
      ISBN = {},
}

@Article{[20], 
  author = 	 {E. R. E. Denton and  M. Holden and E. Chris and J. M. Jarosz and  D. Russell-Jones and  J. Goodey and T. C. S. Cox and and D. L. G. Hil},
  title = 	 {The identification of cerebral volume changes in treated growth hormonedeficient adults using serial 3D MR image processing},
  journal = 	 {Journalof Computer Assisted Tomography},
  year = 	 {2000},
  volume = 	 {24},
  number = 	 {1},
  pages = 	 {139--145}}

@Article{[21], 
  author = 	 {M. Holden and  E. R. E. Denton and J. M. Jarosz and T. C. S. Cox and C. Studholme and D. J. Hawkes and D. L. G. Hilll},
  title = 	 {Detecting small anatomical changes with 3D serial MR subtraction images},
  journal = 	 {  in Medical Imaging: Image Processing, Proc. SPIE},
  year = 	 {1999},
  volume = 	 {366},
  number = 	 {1},
  pages = 	 {44--55}}

@Article{[22], 
  author = 	 {M. Otte},
  title = 	 {Elastic registration of fMRI data using B´ezier-spline transformations},
  journal = 	 { IEEE Transactions on Medical Imaging},
  year = 	 {2001},
  volume = 	 {20},
  number = 	 {3},
  pages = 	 {193--206}}

@Book{[23], 
  author = 	 {Sa\'eid Homayouni},
  title = 	 {Caract\'erisation des Sc\`enes Urbaines par Analyse des Images Hyperspectrales},
  publisher = 	 { Thesis, Ecole Nationale Sup´erieure des T\'el\'ecommunications de Paris},
  year = 	 {2005},
  address = 	 {}}

@Book{[25], 
  author = 	 {Hyvarinen and E. Karhunen },
  title = 	 {Independent Component Analysis},
  publisher = 	 { },
  year = 	 {2001},
  address = 	 {}}

@Article{[26], 
  author = 	 {Oja and Karhunen J.  and Wang  L.  and  R. Vigario},
  title = 	 {Principal and independent components in neural networks - recent developments},
  journal = 	 {In Proc. VII Italian Workshop Neural Networks WIRN'95, Vietrisul Mare, Italy},
  year = 	 {95},
  pages = 	 {16--35},
  address = 	 {}}

@Article{[27], 
  author = 	 {S. Homayouni },
  title = 	 {Assessment of multi source data fusion methods for improvement of accuracy in urban area classification from remotely sensed data},
  journal = 	 { M\'emoire de D.E.A.},
pages={},
  year = 	 {1998},
  address = 	 {Tarbiat Modaress Tehran, Iran}}

@Article{[28], 
  author = 	 {Hyvarinen},
  title = 	 {Survey on independent component analysis},
   journal = 	 {Neural Computing Surveys},
 volume=       {2},
  year = 	 {99},
  pages = 	 {94 --128}}

@Article{[29], 
  author = 	 {Comon},
  title = 	 {Independent component analysis  a new concept},
   journal = 	 {Signal Processing},
 volume=       {36},
  pages = 	 {287--314},
  year = 	 {1994},
  address = 	 {}}

@Article{[30], 
  author = 	 {R. Linsker},
  title = 	 {Local sunaptic learning rules suffice to maximize mutual information in a linear network},
   journal = 	 {Neural Computation},
 volume=       {4},
  pages = 	 {691--702},
  year = 	 {1992},
  address = 	 {}}

@Article{[31], 
  author = 	 {H. Barlow},
  title = 	 {A Unifying Information-Theroretic Framework for Independent Component Analysis },
   journal = 	 {International journal of computers and mathematics with applications},
 volume=       {},
  pages = 	 {},
  year = 	 {1961},
  address = 	 {}}

@Article{[32], 
  author = 	 {Te-Won Lee and M.Girolami and A.J. Bell and T.J. Sejnowski},
  title = 	 {A Unifying Information-Theroretic Framework for Independent Component Analysis },
   journal = 	 {International journal of computers and mathematics with applications},
 volume=       {39},
  pages = 	 {1--21},
  year = 	 {2000},
  address = 	 {}}


\end{document}